\documentclass[conference]{IEEEtran}
\IEEEoverridecommandlockouts
\usepackage{cite}
\usepackage{amsmath,amssymb,amsfonts}
\usepackage{algorithmic}
\usepackage{graphicx}
\usepackage{textcomp}
\usepackage{xcolor}
\def\BibTeX{{\rm B\kern-.05em{\sc i\kern-.025em b}\kern-.08em
    T\kern-.1667em\lower.7ex\hbox{E}\kern-.125emX}}
\begin{document}

\title{Optimization of Image Embeddings for Few Shot Learning\\}

\author{
\IEEEauthorblockN{Arvind Srinivasan}
\IEEEauthorblockA{\textit{Dept. of CSE } \\
\textit{PES University}\\
Bangalore, India \\
arvind.srini.8@gmail.com}
\and
\IEEEauthorblockN{Aprameya Bharadwaj}
\IEEEauthorblockA{\textit{Dept. of CSE } \\
\textit{PES University}\\
Bangalore, India \\
aprameya.bharadwaj@gmail.com}
\and
\IEEEauthorblockN{Manasa Sathyan}
\IEEEauthorblockA{\textit{Dept. of ECE } \\
\textit{PES University}\\
Bangalore, India \\
manasasathyan@gmail.com}
\and
\IEEEauthorblockN{S Natarajan}
\IEEEauthorblockA{\textit{Professor, Dept. of CSE } \\
\textit{PES University}\\
Bangalore, India \\
natarajan@pes.edu}

}

\maketitle

\begin{abstract}
In this paper we improve the image embeddings generated in the graph neural network solution for few shot learning. We propose alternate architectures for existing networks such as Inception-Net, U-Net, Attention U-Net, and Squeeze-Net to generate embeddings and increase the accuracy of the models. We improve the quality of embeddings created at the cost of the time taken to generate them. The proposed implementations outperform the existing state of the art methods for 1-shot and 5-shot learning on the Omniglot dataset. The experiments involved a testing set and training set which had no common classes between them. The results for 5-way and 10-way/20-way tests have been tabulated.
\end{abstract}


\section{Introduction}
With the growth of technology and ease of access to good quality cameras, there is a large number of high resolution images being captured today. A considerable portion of these images is being analyzed and studied for tasks like object detection, scene understanding, image classification, etc. 

Considering the image classification problem, in many real world scenarios, there is very little data available for every class. So, an image classifier needs to be capable of working with very few images per class. The high quality of images can be both a boon and a bane for classification models. When these models are deployed on cloud platforms for concurrent use, the network bandwidth required to transfer these high quality images may reduce the turn around time of responses. However, in situations where there are few images available per class, the high resolution of these images is a boon. Every part of the feature-rich image should be analyzed in order to obtain maximum information.

Another issue that is commonly faced in any classification task is the difference in training data and observed data. While training the model, care must be taken to ensure that the distribution of the training data will be as close as possible to the distribution of the data that will be observed in the real world. 

To tackle the above issues, this paper proposes a new Few Shot Learning model for Image Classification. By definition, the training set and testing set have a null intersection in the few shot learning problem statement. The trained model should work on unseen testing data as long as the new data belongs to the same domain as the trained data. 

In this paper, the Few Shot Learning problem has been modeled as Message Passing Neural Network. This has been done before in \cite{garcia} The nodes of the graph are the images to be classified. Once this graph is built, the labeled nodes are enhanced. They are represented by their image embeddings. The image embeddings are obtained by using different techniques - CNN, U-Net, SqueezeNet, and another architecture that is inspired by the Inception Network. The message passing network transmits information from the labeled nodes to the unlabeled nodes. Eventually, all the nodes in the graph will be assigned classes. 

The proposed method was tested on the Omniglot dataset with a 74\%:26\% split in training-testing data. The training set and test set were disjoint. 

\section{Related Work}

Human Learning \cite{Lake} has served as a cornerstone of research in the field of few-shot learning. Several different approaches have been proposed, the most promising being the 'Meta-learning' paradigm. The model is trained over a variety of tasks and then tested on a distribution of them, including potentially unseen tasks. Many popular solutions employ the Metric-based Meta Learning method. Here, the objective is to learn a mapping from images to their embeddings such that images that are similar are closer and images that are from different categories, are far apart.

In the 1990s, Bromley and LeCun introduced a signature verification algorithm that used a novel artificial neural network - Siamese Neural Network  \cite{bromley}. Siamese neural networks are a class of neural network architectures that contain identical twin sub-networks i.e., they have the same configuration with the same parameters and weights. Parameter updating is mirrored across both sub-networks. In situations where we have thousands of classes but only a few image examples per class, these networks are popular. In \cite{koch}, a one shot classification strategy is presented that involves the learning of image representations using Siamese neural networks. These features are then used for the one-shot task without any retraining. Discrimination between similar and different pairs of images was done by calculating the weighted L1 distance between the twin feature vectors. This was combined with a sigmoid function. A cross-entropy objective was chosen to train the network. For each pair of images, a similarity score was evaluated. It is assumed that this trained network will then work well to classify a new example from a novel class during the one shot task.   
In \cite{garcia}, a single layer message-passing iteration resembles a Siamese Neural Network. The model learns image vector representations whose euclidean metric is agreeing with label similarities.

Matching Networks have also proven to be an excellent model for one-shot learning tasks. In \cite{oriol}, the architecture possesses the best of both worlds - positives of parametric and the positives of non-parametric models. They acquire information from novel classes very quickly while being able to satisfactorily generalize from common examples. Meta-learning with memory-augmented neural networks \cite{santoro} greatly influences this work. LSTMs \cite{hochreiter} learn rapidly from sets of data fed in sequentially. In addition, the authors of \cite{oriol} employ ideas of metric learning \cite{roweis} based on features learnt. The set representation for images is also prevalent in the graph neural model proposed in \cite{garcia}. However the main difference between the two implementations is that matching networks encode the support set independently of the target image i.e., the learning mechanism employed by them attends to the same node embeddings always, in contrast to the stacked adjacency learning in \cite{garcia}.

Prototypical Networks \cite{snell} were designed to provide a simpler yet effective approach for few-shot learning. They build upon work done in \cite{oriol} and the meta-learning approach to few-shot learning \cite{ravi}, showcasing a better performance than Matching Networks without the complication of Full Context Embedding (FCE). These networks apply an inductive bias in the form of class prototypes. There exists embeddings in which samples from each class cluster around a single prototypical representation which is simply the mean of the individual samples. The query image is then classified by finding the nearest class prototype.

In the graph neural model proposed in \cite{garcia}, Prototypical Networks  information is combined within each cluster. Each cluster is defined by nodes with similar labels. 
 
\section{Proposed Method}

\subsection{UNet for Image Embeddings}
For the task of Image Segmentation in the field of Biomedical Imaging, UNets \cite{olaf} were proposed. The architecture consisted of two sections - a contracting path and a symmetric expansion path. The contracting section contained multiple blocks, each applying two 3x3 convolution layers and a 2x2 max pooling layer, on an input image. The number of feature maps doubled after every block enabling the architecture to capture context effectively. The expansion section aims to preserve the spatial properties of the image, key to generating the segmented image. The architecture does this by concatenating feature maps from the corresponding contraction section. Each block in this section consisted of two 3x3 convolution layers and an up-sampling layer. To maintain symmetry, the number of feature maps halved after each block. For training, a softmax function was applied on every pixel of the resultant segmented image, followed by a cross entropy loss function.

For our proposed model, we extracted embeddings of size 64 from the end of the contraction section by using a fully connected layer. These were then fed into the graph neural model. As stated in \cite{olaf}, UNet's speed is one of its major advantages. In our experiment, the UNet model converged the quickest.

Experimentation was also done by replacing the convolution layers with an augmented convolution layer. Augmented convolution operations concatenate the feature maps produced by normal convolution with feature maps produced to self-attention. For the self attention, the input image is flattened and multihead attention is performed \cite{att}. The feature maps are the output for each head and all these feature maps are then concatenated. This attention mechanism makes use of both feature and spatial subspaces.The concatenation of the traditional feature maps with the attention feature maps is as follows: 
The output of each attention head is given by:
\begin{equation}
    O_h = SOFTMAX(\frac{(XW_q)(XW_k)^T}{\sqrt{d_k^h}})(XW_v)
\end{equation}
Where linear transformations map input X to queries Q, keys K and values V as explained in the paper. The W terms are weights which can be learnt.
\begin{equation}
    Q = XW_q 
\end{equation}
\begin{equation}
    K = XW_k
\end{equation}
\begin{equation}
    V = XW_V
\end{equation}

The outputs of all the heads are then concatenated as:
\begin{equation}
 MHA(X) = Concat[O_1,.,.,...O_n]W^o
\end{equation}
Using this, the attention augmented convolution is designed as follows:
\begin{equation}
 AAConv(X) = Concat[Conv(X),MHA(X)]
\end{equation}

Where Conv(X) are the traditional convolutional feature maps. However, after implementing the attention augmented convolution, the following points were noted. 1) When only the first convolutional layer was replaced with the AAConv layer, the training accuracy improved but however, the test accuracy was the same as our UNet implementation. However, the running time of this network was much slower than the UNet with traditional convolution layers. 2) When the first two conv layers were replaced with AAConv layers, the network became even more slower and even the test accuracy reduced. 3) When all three convolution layers were replaced with AAConv layers, the test set accuracy reduced by about 3\% and the network was painfully slow. Due to these reasons, AAConv layers were not incorporated in further experiments.

\begin{figure}[]
\centerline{\includegraphics[scale=0.6]{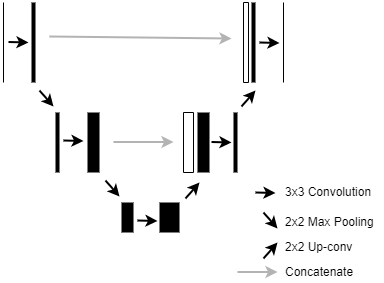}}
\caption{UNet Architecture}
\label{fig}
\end{figure}

\begin{figure}[]
\centerline{\includegraphics[scale=0.6]{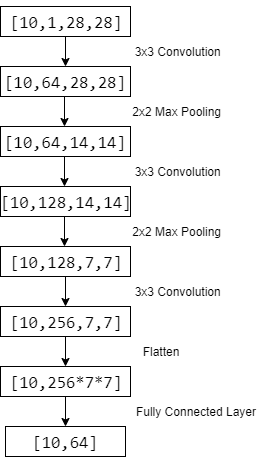}}
\caption{UNet Contraction Section for Image Embeddings}
\label{fig}
\end{figure}

\subsection{Attention U-Net for Image Embeddings}
To increase the performance in the field of image segmentation, the Attention U-Net \cite{attu} was created. The paper proposes a novel gate for attention for medical data. Two feature maps with individual 1x1x1 convolutions are added and then the ReLU function is applied on them. After this, 1x1x1 convolutions is performed again and passed through the sigmoid function. After this step, it goes through resampler. This makes the feature maps same as the ones to be multiplied with. Now concatenation is done with feature maps which are upsampled from the lower level. This gate was incorporated to the UNET architecture as shown in the diagram. A fully connected layer was added after the final encoding layer to extract the image embeddings from this network. However, this network did not perform very well. One reason for this might be the datasets used for the image segmentation task have very large dimensions. Hence, in those images it makes sense to use attention gates to obtain features from different regions of the image. However, the dataset used here consists of 28x28 images. Hence using attention gates might not have any value here. This also supported  by the fact that adding AAConv layers also did not help.
 \begin{figure}[]
\centerline{\includegraphics[scale=0.4]{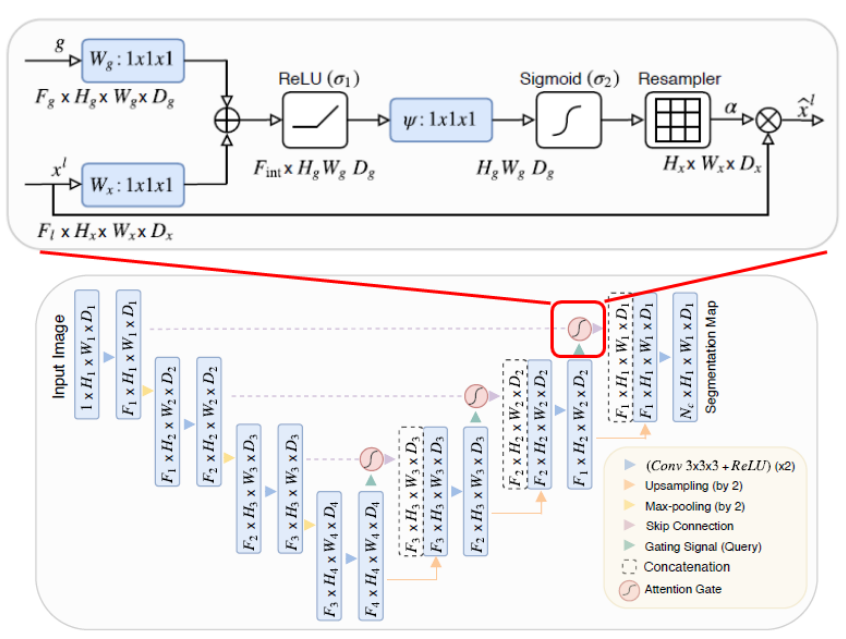}}
\caption{The Attention Gate and Attention U-Net Architecture}
\label{fig}
\end{figure}

\subsection{SqueezeNet for Image Embeddings}

SqueezeNet \cite{squeezenet} was invented to reduce the number of computations that are made to generate image embeddings for classification. The key design choices were to replace 3x3 filters with 1x1 filters to reduce number of parameters by 9 times, decrease the number of input channels to 3x3 filters, and to delay downsampling to retain maximum number of features till the end. While the first two strategies focus on computational efficiency, the third strategy is to maximize the accuracy of the embeddings.

To realize this architecture, the fire module was introduced. The fire module consists of a squeeze phase and an expand phase. The squeeze phase consists of ’s’ 1x1 convolutional filters. The expand phase comprises ‘r’ 1x1 and 3x3 convolutional filters. 

In our experiments with SqueezeNet, we varied the s/r hyper parameter. Like suggested in \cite{squeezenet}, the s/r ratio of 0.125 performed the fastest. An s/r ratio of 0.75 with an equal split of 1x1 and 3x3 filters in the expand phase provided the highest classification accuracy.

Variations with simple and complex bypasses in the network were also experimented with. The simple bypass was found to perform the best.

\begin{figure}[]
\centerline{\includegraphics[scale=0.6]{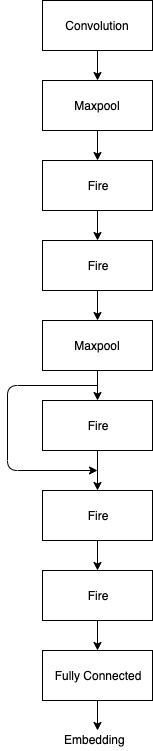}}
\caption{SqueezeNet Architecture for Image Embeddings}
\label{fig}
\end{figure}

\subsection{Modified InceptionNet for Image Embeddings}
Prior to the discovery of the Inception Network, it was thought that as networks go deeper, they get better. The Inception network employs multiple filters at the same level and hence results in a wider network. The inception network uses factorized convolutions to improve computational speed. 
A 5x5 convolution can be filtered into two 3x3 convolutions to speed up the network. The traditional Inception net is used for image classification. To perform this task, it employs to auxiliary classifiers which perform as regularizers to the network. However, the task in hand is to generate image embeddings. To do this, the network architecture was as follows:
Only one of the 5x5 were factorized to two 3x3 as even though it leads to speedup, it may result in loss of information. After concatenation, the dimensions obtained are 10x256x7x7. This was then flattened to 10x12544. After this the fully connected layer is added which reduces the dimensions to 10x64. The auxiliary classifiers were removed and batch normalization is added to the fully connected layer. After this, the vectors are passed through the leaky relu function. The vectors are the embeddings of 10 images and have the dimension of 10x64.

\begin{figure}[]
\centerline{\includegraphics[scale=0.50]{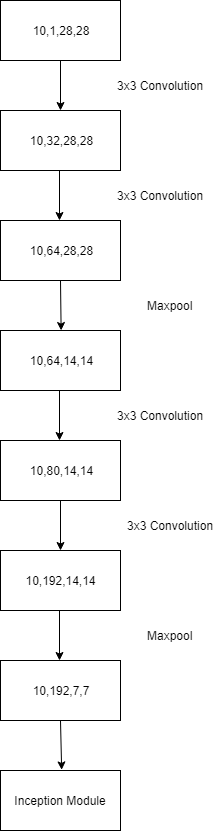}}
\caption{ Architecture Before the Inception Module}
\label{fig}
\end{figure}

\begin{figure}[]
\centerline{\includegraphics[scale=0.5]{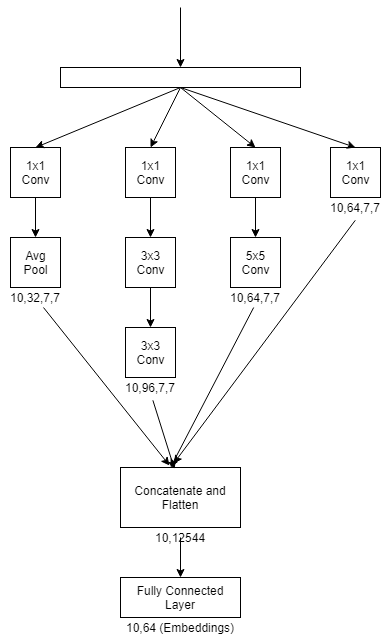}}
\caption{Inception Module}
\label{fig}
\end{figure}

\subsection{R2U-Net for Image Embeddings}
This network proposed by \cite{r2u} integrates the traditional CNN operations with recurrent and residual operations to be used in the U-Net architecture. If x$_{l}$ is the input to the l$^{th}$ layer and (i,j) represents a pixel's location in the k$_{th}$ feature map, then:
\begin{equation}
    O^{l}_{ijk}(t) =  (w_{k}^f)^{T}*x_{l}^{f(i,j)}(t) +(w_{k}^r)^{T}*x_{l}^{r(i,j)}(t-1) + b_k
\end{equation}
Here w$_k^f$ and w$_k^r$ are the weights of the standard convolutional layer and b$_k$ is the bias. The outputs from this are fed into the ReLU function.
\begin{equation}
    F(w_l,x_l) = max(0,O^{l}_{ijk}(t))
\end{equation}
 For this to be used as the R2U-Net, the final outputs are passed through the residual block. 
 \begin{equation}
     x_(l+1) = x_l + F(w_l,x_l)
 \end{equation}
 The x$_{l+1}$ term is used as input for the up-sampling or sub-sampling in the encoder units. After the final encoder layer, a fully connected layer is used to obtain the image embeddings. 
 
 \begin{figure}[]
\centerline{\includegraphics[scale=0.6]{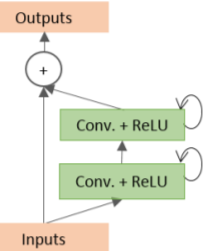}}
\caption{Recurrent Residual Convolution Unit}
\label{fig}
\end{figure}

\begin{figure}[]
\centerline{\includegraphics[scale=0.6]{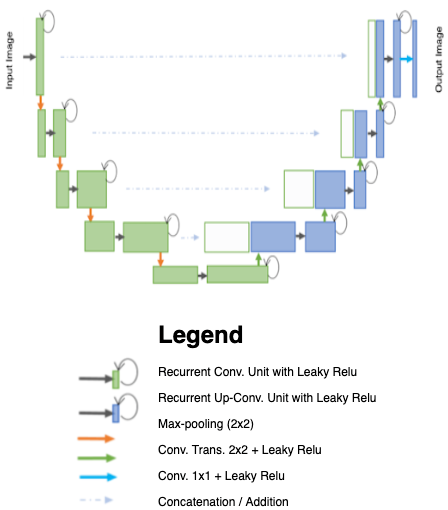}}
\caption{R2U-Net Architecture}
\label{fig}
\end{figure}

\section{Datasets}

The performance of our model was tested on a standard benchmark for one shot learning techniques - Omniglot dataset \cite{omniglot}. It contains 1623 characters from 50 different alphabets. As done in \cite{oriol}, the dataset was split into a training set of 1200 class and a testing set of 423 classes. The dataset was also augmented by rotating the images by multiples of 90 degrees as suggested in \cite{santoro}.

The mini-imagenet dataset was experimented with but later dropped due its enormous memory requirements that exceeded the specifications of the systems used to run the experiments.

\begin{table}[htbp]
\caption{1-shot learning accuracies}
\begin{center}
\begin{tabular}{|c|c|c|}
\hline
\textbf{Model} & \textbf{5-Way}& \textbf{20-Way} \\
\hline
Graph Neural Network \cite{garcia} & 97.81\% & 93.50\% \\
\hline
Proposed Method with U-Net & 98.08\% & 95.53\% \\
\hline
Proposed Method with Inception Net & 98.31\% & 95.21\% \\
\hline
Proposed Method with Attention U-Net & 96.83\% & 93.77\% \\
\hline
Proposed Method with Squeeze Net & 94.54\% & 92.64\% \\
\hline
\end{tabular}
\label{tab1}
\end{center}
\end{table}

\begin{table}[htbp]
\caption{5-shot learning accuracies}
\begin{center}
\begin{tabular}{|c|c|c|}
\hline
\textbf{Model} & \textbf{5-Way}& \textbf{10-Way} \\
\hline
Graph Neural Network \cite{garcia} & 99.32\% & 98.93\% \\
\hline
Proposed Method with U-Net & 99.48\% & 99.21\% \\
\hline
Proposed Method with Inception Net & 99.31\% & 98.97\% \\
\hline
Proposed Method with Attention U-Net & 98.73\% & 97.67\% \\
\hline
Proposed Method with Squeeze Net & 98.26\% & 97.44\% \\
\hline
\end{tabular}
\label{tab1}
\end{center}
\end{table}

\section{Results and Conclusions}
The experiments for this study were run on a system with 20 Intel(R) Xeon(R) ES-2640s. The system also had an Nvidia GeForce GTX1080 with 8 GB of RAM.

From table 1 it is seen that for the tasks of 1-shot 5-way learning and 1-shot 20-way learning, the proposed implementations of the U-Net and Inception-Net to generate embeddings outperformed the network proposed by \cite{garcia} in our working environment. Their 1-shot 5-way and 1-shot 20-way learning models are the state of the art for few shot learning on the Omniglot dataset. 

In the 5-shot learning tasks, the proposed U-Net architecture outperforms the method proposed in \cite{garcia}. The results of the Inception Network architecture are similar to the results of the original work. 

In the case of Attention U-Net, although the model performed well, it did not outperform the proposed U-Net model. This could be due to the fact that the omniglot dataset consists of 28x28-dimensional images. Using the attention mechanism is a waste of computation as the model is required to learn many more unnecessary features. Coming to the case of the R2U-Net, though we hoped that this network would perform the best, it did not run as the memory requirements exceeded available RAM on the system. However, this model might perform well given sufficient hardware specifications.

\vspace{12pt}


\begin{thebibliography}{00}
\bibitem{garcia} Garcia, Victor, and Joan Bruna. "Few-shot learning with graph neural networks." arXiv preprint arXiv:1711.04043 (2017).
\bibitem{Lake} Lake, Brenden M., Ruslan Salakhutdinov, and Joshua B. Tenenbaum. "Human-level concept learning through probabilistic program induction." Science 350, no. 6266 (2015): 1332-1338.
\bibitem{bromley} Bromley, Jane, Isabelle Guyon, Yann LeCun, Eduard Säckinger, and Roopak Shah. "Signature verification using a" siamese" time delay neural network." In Advances in neural information processing systems, pp. 737-744. 1994.
\bibitem{koch} Koch, Gregory, Richard Zemel, and Ruslan Salakhutdinov. "Siamese neural networks for one-shot image recognition." In ICML deep learning workshop, vol. 2. 2015.
\bibitem{oriol} Vinyals, Oriol, Charles Blundell, Timothy Lillicrap, and Daan Wierstra. "Matching networks for one shot learning." In Advances in neural information processing systems, pp. 3630-3638. 2016.
\bibitem{santoro} Santoro, Adam, Sergey Bartunov, Matthew Botvinick, Daan Wierstra, and Timothy Lillicrap. "Meta-learning with memory-augmented neural networks." In International conference on machine learning, pp. 1842-1850. 2016.
\bibitem{hochreiter} Hochreiter, Sepp, and Jürgen Schmidhuber. "Long short-term memory." Neural computation 9, no. 8 (1997): 1735-1780.
\bibitem{roweis} Roweis, Sam, Geoffrey Hinton, and Ruslan Salakhutdinov. "Neighbourhood component analysis." Adv. Neural Inf. Process. Syst.(NIPS) 17 (2004): 513-520.
\bibitem{snell} Snell, Jake, Kevin Swersky, and Richard Zemel. "Prototypical networks for few-shot learning." In Advances in Neural Information Processing Systems, pp. 4077-4087. 2017.
\bibitem{ravi} Ravi, Sachin, and Hugo Larochelle. "Optimization as a model for few-shot learning." (2016).
\bibitem{olaf} Ronneberger, Olaf, Philipp Fischer, and Thomas Brox. "U-net: Convolutional networks for biomedical image segmentation." In International Conference on Medical image computing and computer-assisted intervention, pp. 234-241. Springer, Cham, 2015.
\bibitem{alom} Alom, Md Zahangir, Mahmudul Hasan, Chris Yakopcic, Tarek M. Taha, and Vijayan K. Asari. "Recurrent residual convolutional neural network based on u-net (R2U-net) for medical image segmentation." arXiv preprint arXiv:1802.06955 (2018).
\bibitem{oktay} Oktay, Ozan, Jo Schlemper, Loic Le Folgoc, Matthew Lee, Mattias Heinrich, Kazunari Misawa, Kensaku Mori et al. "Attention u-net: Learning where to look for the pancreas." arXiv preprint arXiv:1804.03999 (2018).
\bibitem{squeezenet} Iandola, Forrest N., Song Han, Matthew W. Moskewicz, Khalid Ashraf, William J. Dally, and Kurt Keutzer. "SqueezeNet: AlexNet-level accuracy with 50x fewer parameters and< 0.5 MB model size." arXiv preprint arXiv:1602.07360 (2016).
\bibitem{att}Vaswani, Ashish, Noam Shazeer, Niki Parmar, Jakob Uszkoreit, Llion Jones, Aidan N. Gomez, Łukasz Kaiser, and Illia Polosukhin. "Attention is all you need." In Advances in neural information processing systems, pp. 5998-6008. 2017.
\bibitem{attu} Oktay, Ozan, Jo Schlemper, Loic Le Folgoc, Matthew Lee, Mattias Heinrich, Kazunari Misawa, Kensaku Mori et al. "Attention u-net: Learning where to look for the pancreas." arXiv preprint arXiv:1804.03999 (2018).
\bibitem{r2u}Alom, Md Zahangir, Mahmudul Hasan, Chris Yakopcic, Tarek M. Taha, and Vijayan K. Asari. "Recurrent residual convolutional neural network based on u-net (R2U-net) for medical image segmentation." arXiv preprint arXiv:1802.06955 (2018).
\bibitem{omniglot}Lake, Brenden M., Ruslan Salakhutdinov, and Joshua B. Tenenbaum. "Human-level concept learning through probabilistic program induction." Science 350, no. 6266 (2015): 1332-1338.
\end{thebibliography}
\end{document}